\documentclass[conference]{IEEEtran}
\IEEEoverridecommandlockouts
\usepackage{lipsum}
\usepackage[numbers,sort&compress]{natbib}
\usepackage{amsmath,amssymb,amsfonts}
\usepackage{algorithm}
\usepackage{algpseudocode}
\usepackage{graphicx}
\usepackage{textcomp}
\usepackage[dvipsnames]{xcolor}
\usepackage[pagebackref,breaklinks,colorlinks]{hyperref}
\usepackage[nameinlink]{cleveref}
\usepackage{subcaption}
\usepackage{mathrsfs}
\usepackage{booktabs}
\usepackage{colortbl}
\usepackage[normalem]{ulem}
\usepackage{cuted}

\newcommand{\ie}{i.e.\ }
\newcommand{\eg}{e.g.\ }
\newcommand{\cf}{cf.\ }

\def\BibTeX{{\rm B\kern-.05em{\sc i\kern-.025em b}\kern-.08em
    T\kern-.1667em\lower.7ex\hbox{E}\kern-.125emX}}
    
\begin{document}

\title{
    Detecting Pose Estimation Failures\\ via Keypoint Self-Consistency \\
}

\author{
    \IEEEauthorblockN{Robin Chan$^*$\thanks{$^*$Work done while visiting researcher in CVLab at EPFL in Lausanne.}}
    \IEEEauthorblockA{
        \textit{Mathematical Modeling of Industrial Life Cycles} \\
        \textit{Institute of Mathematics, Technische Universität Berlin}\\
        Email: \href{mailto:chan@math.tu-berlin.de}{chan@math.tu-berlin.de}
    }
    \vspace{-4em}
}

\pagestyle{plain}
\maketitle

\begin{strip}
    \centering
    \includegraphics[width=0.87\linewidth, trim=0 4.8cm 0 1cm, clip]{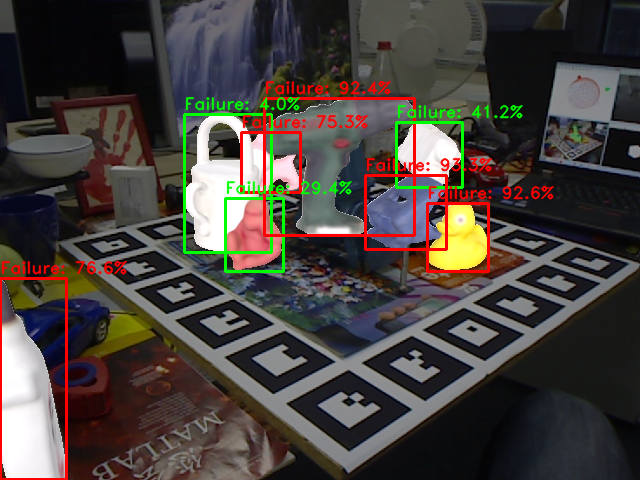}
    \vspace{.5em}
    \captionof{figure}{Pose estimation failure detection results on LINEMOD Occluded \cite{hodan2018bop}. The objects are rendered with their estimated poses. The corresponding bounding boxes are color-coded by rotation error: green indicates a correct pose and red an incorrect pose. The predicted failure probability from our failure detection model is annotated on each bounding box.}
    \label{fig:failure-detection}
\end{strip}

\begin{abstract}
    One common approach to pose estimation involves predicting object keypoints in an image, followed by using Perspective-n-Point algorithms to compute the object's rotation and translation relative to the camera. While rotations preserve object shapes, this property is often neglected in keypoint-based pose estimation methods, where keypoints are typically predicted independently from each other. As imprecise keypoint predictions negatively affects pose estimation accuracy, it also limits its reliability in downstream tasks. In this work, we explore whether such inaccurate pose estimates can be identified by simply examining spatial locations between 2D keypoints. We propose a set of hand-crafted geometric features that capture the self-consistency of keypoint predictions, including pairwise distances, reprojection consistency, as well as render and mask consistency. Despite its simplicity, a logistic regression classifier trained on these features reliably detects pose estimation failures, outperforming confidence-based approaches like conformal keypoint predictions that rely solely on keypoint uncertainty.\\
    \indent Code available: \href{https://github.com/robin-chan/meta-pose}{https://github.com/robin-chan/meta-pose}\\
\end{abstract}


\section{Introduction}\vspace{1em}

6D object pose estimation is the computer vision task that deals with determining the orientation and translation of an object displayed in an image relative to the camera capturing the image. Recent advancements have shown impressive success \cite{labbe2023megapose, wen2024foundationpose, Lee2025any}. In general, the 6D pose of an object can be described by means of a $3 \times 3$ rotation matrix and a $3$ dimensional translation vector, representing the regression targets.

Although these parameters can be regressed directly \cite{kendall2015posenet, xiang2017posecnn}, more recent approaches favor two-stage methods for better accuracy. Such two-stage approaches first localize predefined object keypoints in the image and then solve a Perspective-n-Point (PnP) problem \cite{pavlakos2017semantic, peng2019pvnet, Chen2020End, schmeckpeper2022semantic, yang2023conformal} to obtain the 6D pose.
Typically, only the detection of these 2D keypoints in the image involves deep learning. Their locations can be predicted via intermediate probabilistic heatmaps \cite{pavlakos2017semantic, oberweger2018making, schmeckpeper2022semantic, yang2023conformal}.

In most of these 2D keypoint prediction variants, however, the keypoints are treated independently. They do not account for their spatial correlation of the keypoints in the image, which naturally exist since they represent parts of an object with a fixed geometric relationship. While the relative positions of keypoints of a rigid 3D object remain consistent under rotation in 3D space, this is not the case after projecting them to 2D plane. The correlation still persists but the relative positions change depending on the rotation.

In this work, we investigate whether pose estimation failures can be detected purely from predicted 2D keypoint locations, without requiring additional sensors beyond the monocular camera input available in the standard pose estimation pipeline, or computationally expensive learning based components. To this end, we design hand-crafted geometric features that encode the self-consistency of predicted keypoints with respect to the known 3D structure of the object. Concretely, we construct features from pairwise keypoint distances, including keypoint predictions from the input image, reprojections under the estimated pose, or predictions on the rendered object. We demonstrate that a lightweight logistic regression classifier trained on these features reliably detects pose estimation failures, requiring only a small calibration set drawn from the target domain. Thereby, our approach improves the robustness of the 6D pose estimation pipeline and subsequent downstream tasks.\\

Our contributions are summarized as follows:
\begin{itemize}
    \item We propose \textbf{Meta Pose}, a lightweight failure detection framework for keypoint-based 6D pose estimation that requires no additional sensors or costly learning-based components beyond a standard pose estimation pipeline.
    \item We introduce a set of hand-crafted geometric features encoding the self-consistency of keypoint predictions across complementary sources, including pairwise distances, reprojection, as well as render and mask consistency. Combined with a logistic regression, these features provide reliable and well-calibrated failure probabilities.
    \item We demonstrate on the challenging LINEMOD Occluded dataset that our approach consistently outperforms established baselines, such as conformal keypoint prediction, across multiple evaluation metrics despite its simplicity, which highlights the value of geometric self-consistency as a signal for pose estimation reliability.
    \item We show that a lightweight, render-free subset of our hand-crafted features is sufficient to match the performance of the full feature set at a fraction of the computational cost, indicating that reliable failure detection does not require access to rendering or segmentation beyond the standard pose estimation pipeline.
\end{itemize}

The remainder of this paper is organized as follows. In \Cref{sec:related-work}, we review related work. In \Cref{sec:methodology}, we detail the proposed geometric consistency features and the pose estimation failure detection framework. In \Cref{sec:setup}, we describe the experimental setup, including the dataset, evaluation protocol, and baselines. In \Cref{sec:results}, we present both qualitative and quantitative results. Finally, \Cref{sec:ablation} concludes with an ablation study.

\section{Related Work}\label{sec:related-work}\vspace{1em}

\textbf{6D Pose Estimation from Images.}
Early approaches to 6D pose estimation regress rotation and translation in a single forward pass \cite{kendall2015posenet, xiang2017posecnn, Wang2021GDRnet}, enabling real-time inference. However, the non-Euclidean structure of the rotation space makes direct regression problematic, for which alternative rotation representations have been proposed to mitigate the issues \cite{zhou2019continuity, labbe2020cosypose}. These methods rely primarily on image features and lack geometric reasoning, which is why they tend to struggle under occlusion and clutter.

More favorable are two-stage approaches that first establish 2D--3D correspondences between image and 3D object to then recover the pose by solving a Perspective-n-Point (PnP) problem \cite{lu2018review, Lepetit2009EPnP}. Classical methods derive these correspondences from hand-crafted features \cite{lowe1999object, rothganger20063d}. Learning-based methods instead predict the 2D locations of predefined 3D object keypoints via direct regression \cite{rad2017bb8, kehl2017ssd, tekin2018real}, pixel-wise voting \cite{peng2019pvnet, hu2019segmentation}, or probabilistic heatmaps \cite{pavlakos2017semantic, oberweger2018making, schmeckpeper2022semantic, yang2023conformal}. In all of these variants, keypoints are typically predicted independently, without explicitly enforcing the geometric constraints that arise from the rigid 3D structure of the object. Consequently, localization errors in individual keypoints may propagate directly to the pose recovered by the PnP solver. Inconsistent keypoint predictions can therefore lead to significant pose estimation failures. This motivates the need for reliable failure detection as a component of the two-stage pipeline.\\

\textbf{Failure Detection in 6D Pose Estimation.}
The problem of detecting when a model's prediction is incorrect has been studied in computer vision and can be tied naturally to the problem of uncertainty quantification. One established early baseline in image classification thresholds the neural network's maximum softmax confidence \cite{hendrycks2017baseline}. Subsequent work has refined this idea by learning confidence predictors \cite{corbiere2019addressing, devries2018learning}, applying input preprocessing to sharpen the confidence gap \cite{liang2017enhancing}, or exploiting feature-based statistics \cite{lee2018simple}. These ideas have been extended to more complex computer vision tasks such as semantic segmentation by aggregating statistics over softmax distributions in ensemble like fashion \cite{rottmann2020prediction, chan2021entropy}. Here in this work we use ensembles of features to predict whether a pose is correct or incorrect.

In the specific context of 6D pose estimation, failure detection remains relatively understudied. \citet{richter2019towards} introduced the idea of including uncertainty from an image encoding CNN directly in the PnP solver to predict the model's own failures. \citet{shi2021fast} and \citet{wursthorn2024uq} explore ensemble-based uncertainty quantification, applying deep ensembles to multi-stage pose estimators and proposing metrics to quantify the quality of pose estimates. Other methods explicitly model orientation uncertainty by learning distributions over orientations \cite{Okorn2020LearningOD} or by predicting dense correspondence distributions \cite{haugaard2022surfemb}. Another line of strategies selects a pose among a set of pose hypotheses based on depth or visual feature consistency \cite{Meng2020, Brachmann2016uncertainty, lin2022keypoint}.

Closer to our goal of detecting incorrect pose estimates without ground truth is the work by \citet{Quentin2024}. They propose a render-based uncertainty score that compares the estimated pose against the corresponding instance segmentation mask, operating at the level of silhouettes rather than keypoints. Failure detection has also been tackled in the work by \citet{schneider2026tracing}, where pose estimation pipeline is decomposed into failure detection, error source attribution, and targeted mitigation. A small classifier can be trained to distinguish between error types for ICP-based pose estimation.

An emerging line of work addresses the gap of missing statistical guarantees via conformal prediction \cite{vovk2005algorithmic, papadopoulos2002inductive}. \citet{yang2023conformal} apply inductive conformal prediction to construct keypoint prediction sets with coverage guarantees and propagate these geometrically to obtain worst-case pose error bounds. \citet{Shaikewitz2025PoseUncertaintySets} extend this direction to visual pose uncertainty bounds around objects. While these methods provide coverage guarantees on individual keypoints, such guarantees do not transfer directly to the final pose, which depends jointly on all keypoint predictions. Furthermore, their practical utility for binary failure detection, \ie predicting whether a pose estimate exceeds a specific rotation error threshold, has not been explicitly investigated. Our work addresses this gap by training a lightweight logistic regression model on hand-crafted geometric features encoding self-consistency of keypoints, directly targeting failure detection of pose estimation without additional sensors beyond the standard monocular camera input.\\

\section{Methodology}\label{sec:methodology}\vspace{1em}

We consider the task of keypoint-based 6D pose estimation, where the pose is defined by the rotation matrix $\mathbf{R} \in \mathrm{SO}(3)$ and the translation vector $\mathbf{t}\in\mathbb{R}^3$. Given a set of predefined 3D object keypoints $\mathcal{V} \subset \mathbb{R}^3$ and camera intrinsics $\mathbf{K}$, the corresponding 2D projections are obtained as
\begin{equation}
    \mathcal{X} = \{ \pi(\mathbf{K}, \mathbf{R}, \mathbf{t}, \mathbf{v}) \mid \mathbf{v} \in \mathcal{V} \} \subset \mathbb{R}^2,
\end{equation}
where $\pi(\cdot)$ denotes the projection function. Keypoint-based methods first predict the 2D locations $\hat{\mathcal{X}}^\text{im}$ of the keypoints $\mathcal{X}$ in the image, from which the rotation and translation is recovered deterministically via a PnP algorithm given 2D--3D point correspondences between $\hat{\mathcal{X}}^\text{im}$ and $\mathcal{V}$.

\subsection{Rotation Error}
Let $\hat{\mathbf{R}}$ and $\mathbf{R}$ denote two rotation matrices corresponding to the estimated rotation and ground truth rotation, respectively. The geodesic distance between them is measured by the rotation angle (in degrees)
\begin{equation}\label{eq:geo-dist}
    \theta = \arccos \left( \frac{\mathrm{tr}(\hat{\mathbf{R}} \mathbf{R}^\top) - 1}{2} \right)\cdot \frac{180}{\pi} \in [0,180],
\end{equation}
where $\hat{\mathbf{R}} \mathbf{R}^\top \in \mathrm{SO}(3)$ is the relative rotation.
A pose estimate is called \emph{correct} if $\theta(\hat{\mathbf{R}}, \mathbf{R}) \leq \tau$ for a given threshold $\tau \in [0,180]$, and \emph{incorrect} otherwise.

\subsection{Feature Construction}
We construct a feature vector for each pose estimate. With these feature vectors, we aim at capturing geometric consistency of predicted keypoints, without requiring ground truth annotations at test time. Since all keypoints are typically predicted independently, deviations from the expected object structure serves as indicator for inaccurate predictions and, consequently, incorrect pose estimates. All features are constructed solely from keypoint locations, making the approach computationally lightweight without requiring processing steps beyond the pose estimation pipeline itself.
In what follows, we discuss the feature construction for only one object in one image for simplicity.\\

\textbf{Relative Distances.} Let $\hat{\mathbf{x}}_k^\text{im} \in \hat{\mathcal{X}}^\text{im} \subset \mathbb{N}^2$ denote the 2D location of the $k$-th keypoint obtained by the keypoint detector given an original input image. To capture the geometric structure of the 
predicted keypoints, we compute pairwise distances between all predicted image keypoints,
\begin{equation}
    \mathbf{d}^\text{rel} = \hat{t}_z \cdot \left( \| \hat{\mathbf{x}}_i^\text{im} - \hat{\mathbf{x}}_j^\text{im} \|_2  \right)_{i,j=1, i \neq j}^{|\mathcal{V}|} \in \mathbb{R}^{|\mathcal{V}|^2-|\mathcal{V}|}~,
\end{equation}
where the estimated distance $\hat{t}_z \in \mathbb{R}_+$ of the object from the camera is used for scaling.
Deviations in pairwise keypoint distances from those expected under a correct pose indicate inconsistent keypoint predictions.\\

\textbf{Reprojection Consistency.} Let $\hat{\mathbf{x}}_k^\text{repr} \in \hat{\mathcal{X}}^\text{repr} \subset \mathbb{R}^2$ denote the 2D location of the $k$-th keypoint obtained by projecting the 3D keypoints $\mathcal{V}$ back to the image plane using the estimated rotation $\hat{\mathbf{R}}$ and translation $\hat{\mathbf{t}}$ after solving the PnP problem, \ie $\hat{\mathcal{X}}^\text{repr} = \{ \pi(\mathbf{K}, \hat{\mathbf{R}}, \hat{\mathbf{t}}, \mathbf{v}) \mid \mathbf{v} \in \mathcal{V} \}$. We measure the consistency between predicted and reprojected keypoints as
\begin{equation}
    \mathbf{d}^\text{repr} = \hat{t}_z \cdot\left( \| \hat{\mathbf{x}}_i^\text{im} - \hat{\mathbf{x}}_i^\text{repr} \|_2 \right)_{i=1}^{|\mathcal{V}|} \in \mathbb{R}^{|\mathcal{V}|}~.
\end{equation}
A consistent pose estimate should yield small reprojection distances from all predicted keypoints.\\

\begin{figure}[t]
    \centering
    \captionsetup[subfigure]{labelformat=empty, justification=centering}
    \subcaptionbox{$\theta=0.98^{\circ}$, $\bar{d}^2 = 1.49$}{\includegraphics[width=0.99\linewidth]{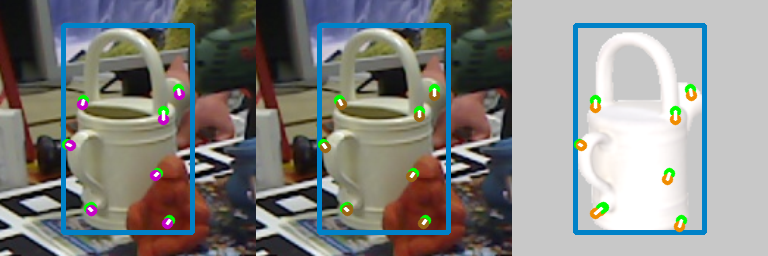}}\\[6pt]
    \subcaptionbox{$\theta=9.48^{\circ}$, $\bar{d}^2 = 14.35$}{\includegraphics[width=0.99\linewidth]{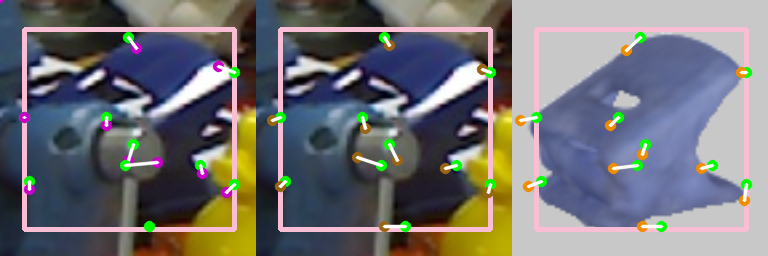}}\hfill
    \caption{Examples showing keypoint consistency features. From left to right: keypoint predictions on original image (pink), reprojected keypoints given pose estimate (brown), keypoint prediction on rendered image given pose estimate (orange). In each image the ground truth points are green with its corresponding prediction indicated by the white line. Moreover, $\bar{d}^2$ quantifies the average spread of all keypoints from their respective centroids.}
    \label{fig:keypoints-consistency}
\end{figure}

\textbf{Render Consistency.} Since the 3D asset is available in keypoint-based pose estimation, we can render the object at the estimated pose and predict keypoint locations on the rendered image, yielding $\hat{\mathcal{X}}^\text{render}$. Let $\hat{\mathbf{x}}_k^\text{render} \in \hat{\mathcal{X}}^\text{render} \subset \mathbb{N}^2$ denote the 2D location of the $k$-th keypoint obtained by the keypoint detector given the rendered image. As the render is generated under controlled conditions without occlusion or motion blur, and as the keypoint detector is typically trained on rendered data, it is expected to perform more reliably on it. We measure the consistency between image and rendered image keypoint predictions as
\begin{equation}
    \mathbf{d}^\text{render} = \hat{t}_z \cdot \left( \| \hat{\mathbf{x}}_i^\text{im} - \hat{\mathbf{x}}_i^\text{render} \|_2 \right)_{i=1}^{|\mathcal{V}|} \in \mathbb{R}^{|\mathcal{V}|}~.
\end{equation}
Discrepancies indicate that the estimated pose given the original image is inconsistent with the estimated pose given the rendered image.\\

\textbf{Centroid Distance.} We compute the centroids for each keypoint across different keypoint predictions. The centroid for the $k$-th keypoint is obtained by
\begin{equation}
    \hat{\mathbf{c}}_k = \frac{1}{3} \left( \hat{\mathbf{x}}_k^\text{im} + \hat{\mathbf{x}}_k^\text{repr} + \hat{\mathbf{x}}_k^\text{render} \right) \in \mathbb{R}^2.
\end{equation}
The distance to the centroid across the keypoints from the different sources
\begin{equation}
    \mathbf{d}^\text{center} = \hat{t}_z \cdot \left( \frac{1}{3} \sum_{\mathbf{x} \in \mathcal{K}_k } \| \mathbf{x} - \hat{\mathbf{c}}_k \|_2 \right)_{k=1}^{|\mathcal{V}|} \in \mathbb{R}^{|\mathcal{V}|},
\end{equation}
where $\mathcal{K}_k = \{\hat{\mathbf{x}}^{\text{im}}_k,\, \hat{\mathbf{x}}^{\text{repr}}_k,\, \hat{\mathbf{x}}^{\text{render}}_k\}$, summarizes the spread of the different keypoint types $\hat{\mathcal{X}}^\text{im}, \hat{\mathcal{X}}^\text{repr}, \hat{\mathcal{X}}^\text{render}$. A consistent pose estimate is expected to cluster all keypoints close to their respective centroids.\\

\textbf{Keypoint Confidence.} A natural measure for the reliability of keypoint predictions is the confidence of the keypoint localization method. Given normalized confidence scores $\mathbf{s} \in [0, 1]^{|\mathcal{V}|}$, the per-keypoint uncertainty is the inverse confidence
\begin{equation}\label{eq:max-uncertainty}
    \mathbf{u} = (1 - s)_{s \in \mathbf{s}} \in [0,1]^{|\mathcal{V}|}~.
\end{equation}
High uncertainty predictions are expected to correlate to incorrect pose estimates.\\

\begin{figure}[t]
    \centering
    \captionsetup[subfigure]{labelformat=empty, justification=centering}
    \subcaptionbox{$d^\text{cham}=0.61$, $r^\text{cov}_{\delta=2} = 0.97$}{\includegraphics[width=0.99\linewidth]{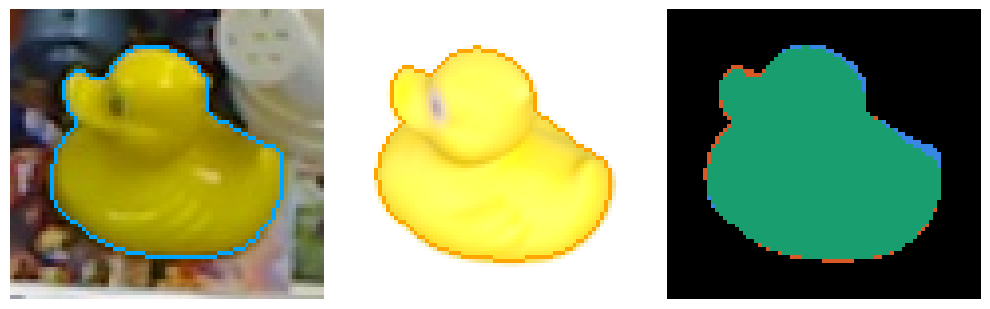}}\\[6pt]
    \subcaptionbox{$d^\text{cham}=1.86$, $r^\text{cov}_{\delta=2} = 0.75$}{\includegraphics[width=0.99\linewidth]{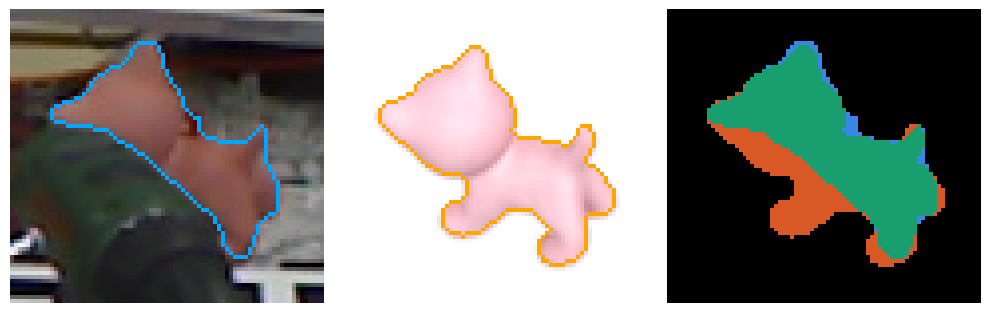}}\hfill
    \caption{Examples showing mask consistency features. From left to right: the segmentation of the object in the original image obtained with SAM \cite{kirillov2023segment} (blue), the rendered object at its estimated pose with its corresponding segmentation (orange), and the overlay of both masks, where agreement is highlighted in green. Below each example we report the Chamfer distance $d^{\text{cham}}$ and coverage ratio $r^\text{cov}_\delta$ with tolerance $\delta=2$.}
    \label{fig:mask-consistency}
\end{figure}

\textbf{Render Mask Consistency.} Similar to \cite{Quentin2024}, we also exploit the segmentation masks to detect pose failures. Let $\mathcal{M}^{\text{render}} \subset \mathbb{Z}^2$ denote the segmentation mask of the rendered object at its estimated pose and $\partial \mathcal{M}^{\text{render}}$ its boundary points. As a prediction of the object's silhouette in the original image, we prompt the Segment Anything Model (SAM)~\cite{kirillov2023segment} with the 2D bounding box used for pose estimation to obtain the segmentation mask and its boundary points $\partial \mathcal{M}^{\text{seg}}$. A pose estimate is expected to be consistent if the rendered silhouette aligns well with the predicted segmentation.

We quantify this alignment using two metrics. 
First, the one directional Chamfer distance between the render and segmentation boundaries is given by
\begin{equation}
    d^{\text{cham}} = \frac{\hat{t}_z}{|\partial \mathcal{M}^{\text{seg}}|}
    \sum_{\mathbf{p} \in \partial \mathcal{M}^{\text{seg}}} \min_{\mathbf{q} \in \partial \mathcal{M}^\text{render}} \|\mathbf{p} - \mathbf{q}\|_2 \in \mathbb{R}_+
    \label{eq:chamfer}
\end{equation}
where a smaller values for $d^{\text{cham}}$ indicates closer agreement between the two silhouettes. This choice
is motivated by better robustness to occlusion since it puts emphasis on visible contour points.

Second, we compute a coverage ratio defined by the fraction of observed segmentation boundary points that lie within a tolerance pixel distance $\delta>0$ of the render boundary
\begin{equation}
    r^\text{cov}_\delta \! = \! \frac{1}{|\partial \mathcal{M}^{\text{seg}}|} \!
     \sum_{\mathbf{p} \in \partial\! \mathcal{M}^{\text{seg}}}
    \!\!\!\!\!\mathbf{1}_{\{
    \min_{\mathbf{q} \in \partial\! \mathcal{M}^{\text{render}}}\! \|\mathbf{p} - \mathbf{q}\|_2 \leq \delta
    \}} \in [0,1]
    \label{eq:coverage}
\end{equation}
where low values indicate complete inconsistency, and high values that the rendered silhouette mostly matches the segmentation boundary.
See \Cref{fig:mask-consistency} for illustration.

\subsection{Pose Estimation Failure Detector}
The above features are computed for each test sample independently. For each test object, this yields an aggregated single full feature vector
\begin{equation}
    \mathbf{f} = \mathrm{concat} \!
    \left(
    \mathbf{d}^\text{rel}\!, \mathbf{d}^\text{repr}\!, \mathbf{d}^\text{render}\!\!, \mathbf{d}^\text{center}\!\!, \mathbf{u}, d^\text{cham}\!, r^\text{cov}_\delta
    \right)  \in \mathbb{R}^{N}
\end{equation}
where $N = 2+\sum_k \dim \mathbf{d}_k$, $\mathbf{d}_k \in \{\mathbf{d}^\text{rel}\!, \mathbf{d}^\text{repr}\!, \mathbf{d}^\text{render}\!, \mathbf{d}^\text{center}\!, \mathbf{u} \}$. Then, $\mathbf{f}$ serves as input to a logistic regression model, which classifies whether the corresponding pose estimate is correct or incorrect, \ie whether $\theta \leq \tau$ or $\theta > \tau$, \cf \Cref{eq:geo-dist}.

Note that other combinations of the presented features are possible as well. We evaluate different combinations in the main experiments in \Cref{sec:results} and ablation study in \Cref{sec:ablation}.

\section{Experiment Setup}\label{sec:setup}\vspace{1em}

In this work, we focus on rotation error. The translation accuracy is largely determined by the quality of the preliminary object localization rather than the pose estimator itself. A sufficiently accurate detection typically yields low translation error. Therefore, pose estimation failure resulting from inaccurate translation is more appropriately addressed at the object detection stage, which is a separate line of research outside the scope of this work. All evaluated method use the same underlying pose estimation pipeline by \cite{schmeckpeper2022semantic}, which localize keypoints from heatmap predictions. Moreover, we use the RANSAC-based implementation by OpenCV \cite{opencv_solvepnp} to solve the Perspective-n-Points (PnP) problems.

\subsection{Evaluation Metrics}
The \textbf{pose error rate} is defined as the fraction of incorrect estimates over the dataset $\mathcal{I}$,
\begin{equation}\label{eq:pose-error}
    \varepsilon_\tau = \frac{1}{|\mathcal{I}|} \sum_{i \in \mathcal{I}} 
    \mathbf{1}_{\{\theta_i > \tau\}}~.
\end{equation}
For increasing error threshold, the pose error rate decreases, \ie it holds $\varepsilon_{\tau_1} \geq \varepsilon_{\tau_2}$ if $\tau_1 \leq \tau_2,~ \tau_1,\tau_2 \in [0,180]$. For the detection of pose estimation failures, this means that the number of positive and negative examples are affected by the chosen error threshold, which in return impact the performance of the failure detection model.

Failure detection is formulated as a binary classification problem, where the positive class corresponds to incorrect pose estimates ($\theta > \tau$) and the negative class to correct ones ($\theta \leq \tau$). We evaluate failure detection performance using four metrics. 

\textbf{Accuracy} measures the overall fraction of correctly classified samples but is sensitive to class imbalance. The \textbf{F$_1$} score, as the harmonic mean of precision and recall, puts emphasis on the detection of failures and is more informative than accuracy under class imbalance. 
Both metrics depend on the choice of a decision threshold. We report results at the threshold which jointly maximizes training accuracy and training F$_1$ score, \ie $t^* = \arg \max_{t \in \mathbb{R}} (\text{Accuracy}(t) + \text{F}_1(t))$.

For threshold-agnostic evaluation of discriminative performance, we report the \textbf{AUROC}, which measures the ability of the classifier to rank pose estimation failures above correct estimates across different decision thresholds. Additionally, we report the \textbf{AUPRC} which is both threshold-agnostic and failure-oriented, making it suited in imbalanced class settings.

\subsection{Evaluation Protocol}
We evaluate the proposed failure detector using a reverse cross-validation protocol. To this end, we partition the 1214 test images of the LINEMOD Occluded dataset~\cite{hodan2018bop} into $5$ folds of (approximately) equal size. The folds are independent, as the images provide uniformly distributed views of objects \cite{hinterstoisser2012model, yang2023conformal}. In each run, one fold ($20\%$ of the data) is used to train the logistic regression classifier for failure detection, while the remaining four folds ($80\%$) serve as test set. This is repeated for each fold, yielding $5$ independent evaluation runs in which every sample is used for training once and evaluated four times in the remaining runs. This protocol is a common evaluation in uncertainty quantification where only a small labeled subset of the target domain is required to calibrate the classifier, while the majority of the data serves as an independent test set \cite{papadopoulos2002inductive}. We additionally report the mean and standard deviation across all $5$ runs to assess the sensitivity of the failure detector to the choice of the data split.

\subsection{Evaluated Methods}

\begin{figure}
    \centering
    \includegraphics[width=.99\linewidth]{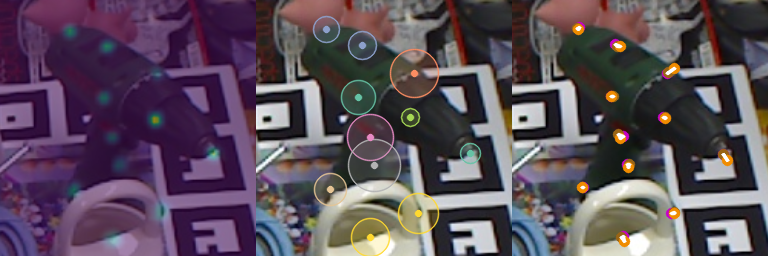}\\[1em]
    \includegraphics[width=.99\linewidth]{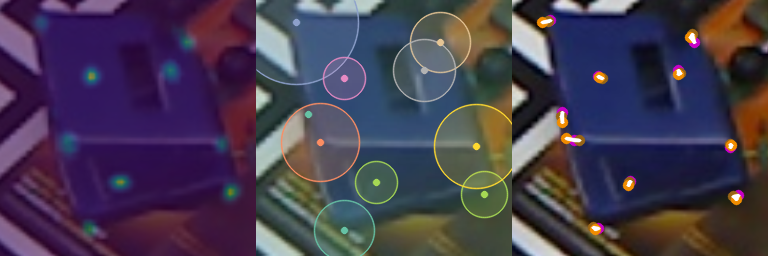}
    \caption{Examples showing different methods. From left to right: keypoint confidence heatmaps, conformal keypoint sets, keypoint consistency features.}
    \label{fig:methods}
\end{figure}

\textbf{Meta Pose Variants.} We consider two variants of our proposed approach, \cf \Cref{sec:methodology}. Here, \textbf{Meta Pose} aggregates only the non-render features, namely $\mathbf{d}^{\text{rel}}$, $\mathbf{d}^{\text{repr}}$, and $\mathbf{u}$. It requires no rendering or additional segmentation model beyond the pose estimation pipeline. \textbf{Meta Pose--Full} additionally includes the render-based features $\mathbf{d}^{\text{render}}$, $\mathbf{d}^{\text{center}}$, $d^{\text{cham}}$, and $r^\text{cov}_{\delta=2}$, which require rendering the object at the estimated pose and, for the mask consistency features, an additional segmentation step. As we observe that the render-based features contribute only marginal gains in failure detection performance while considerably increasing runtime, we adopt the render-free variant as our main approach. Unless stated otherwise, Meta Pose therefore refers to this render-free variant throughout the remainder of the paper. For completeness, we report results for both variants in our numerical results in \Cref{sec:results}, with  accompanying feature ablation in \Cref{sec:ablation}.

\textbf{Maximum Keypoint Confidence Thresholding.} Motivated by the established baseline for failure detection in computer vision introduced by \citet{hendrycks2017baseline}, we evaluate failure detection performance using the keypoint confidence scores produced by the keypoint localization model. The intuition is that correctly estimated poses are expected to arise from accurately localized keypoints, which in turn correspond to high keypoint confidence scores. On the contrary, incorrectly estimated poses are expected to result from poorly localized keypoints with lower keypoint confidences. Following this reasoning, the baseline method aggregates the per-keypoint uncertainties $\mathbf{u}$, as defined in \Cref{eq:max-uncertainty} as the inverse of the keypoint confidence scores, by taking their maximum value. The scalar quantity used for thresholding is thus given by $u^* = \max \mathbf{u}$, where a high value indicates low overall keypoint confidence and therefore signals a pose estimation failure. This baseline method relies solely on the per-keypoint uncertainty of the keypoint detector. It requires no additional computation beyond the keypoint localization model of the standard pose estimation pipeline or calibration, making it the most efficient baseline method for comparison.\\

\textbf{Conformal Keypoint Prediction.}
\citet{yang2023conformal} proposed a conformal keypoint detection framework that predicts sets of possible keypoint locations for each keypoint with statistical guarantees. Specifically, inductive conformal prediction is applied to construct circular regions in the image space based on keypoint localization confidences, each guaranteed to contain the ground truth keypoint location with a chosen coverage probability. Pose uncertainty is then quantified by sampling keypoint locations within the spatial constraints given by the conformal sets, followed by solving a Perspective-n-Point problem for each sample, yielding a distribution over possible poses. From these distribution rotation errors can be estimated by means of statistics of the pose distribution. The intuition is that larger conformal keypoint sets correspond to greater keypoint localization uncertainty and, consequently, higher pose uncertainty. 

While the method has shown to provide reliable keypoint coverage and worst-case error bounds, its use for failure detection has not been explicitly evaluated. We therefore adopt conformal keypoint prediction as our baseline, assuming that pose uncertainty is indicative of incorrect pose estimates. We train a logistic regression classifier on three scalar statistics derived from the distribution of sampled poses, which are the mean, variance, and maximum rotation error between the estimated pose and each sampled pose. This makes the baseline directly comparable to our proposed approach, differing only in the features provided to the logistic regression model as input.

\begin{figure}[t]
    \centering
    \captionsetup[subfigure]{labelformat=empty, justification=centering}
    \subcaptionbox{$\theta=2.16^{\circ}$ \\ $\hat{p} = 29.4\%$}{\includegraphics[width=0.23\linewidth]{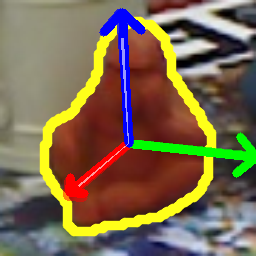}}\hfill
    \subcaptionbox{$\theta=0.98^{\circ}$ \\ $\hat{p} = 4.0\%$}{\includegraphics[width=0.23\linewidth]{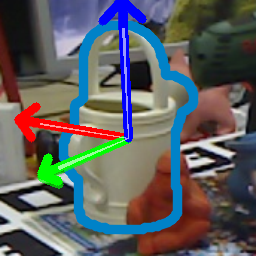}}\hfill
    \subcaptionbox{$\theta=7.92^{\circ}$ \\ $\hat{p} = 75.3\%$}{\includegraphics[width=0.23\linewidth]{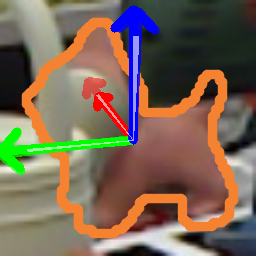}}\hfill
    \subcaptionbox{$\theta=19.04^{\circ}$ \\ $\hat{p} = 92.4\%$}{\includegraphics[width=0.23\linewidth]{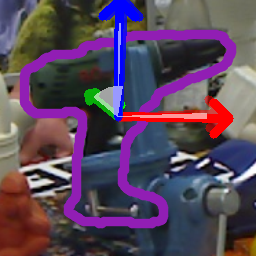}}\\[6pt]
    \subcaptionbox{$\theta=8.40^{\circ}$ \\ $\hat{p} = 92.6\%$}{\includegraphics[width=0.23\linewidth]{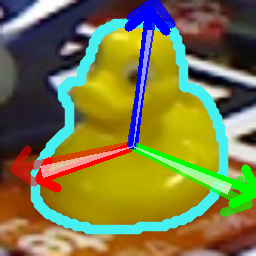}}\hfill
    \subcaptionbox{$\theta=3.66^{\circ}$ \\ $\hat{p} = 41.2\%$}{\includegraphics[width=0.23\linewidth]{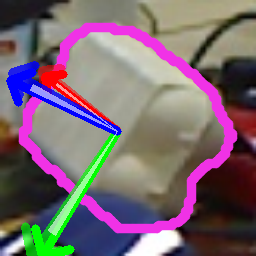}}\hfill
    \subcaptionbox{$\theta=9.92^{\circ}$ \\ $\hat{p} = 76.6\%$}{\includegraphics[width=0.23\linewidth]{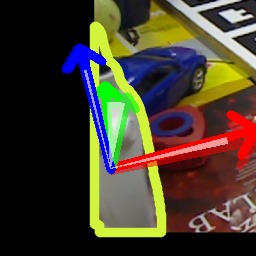}}\hfill
    \subcaptionbox{$\theta=9.48^{\circ}$ \\ $\hat{p} = 93.3\%$}{\includegraphics[width=0.23\linewidth]{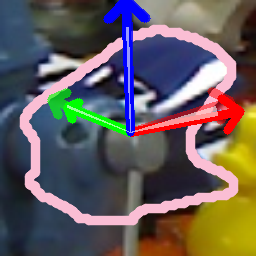}}
\caption{Examples showing estimated and ground truth poses for the different LINEMOD objects \cite{hodan2018bop} in the scene shown in \Cref{fig:failure-detection}. The object contours and solid axes correspond to the estimated pose, transparent axes to the ground truth. Moreover, $\theta$ denotes the rotation error and $\hat{p}$ the our model's probability for a pose estimation failure, \ie $\theta > 5^\circ$.}
\label{fig:grid}
\end{figure}

\section{Numerical Results}\label{sec:results}
\vspace{1em}

\begin{figure*}
    \centering
    \includegraphics[width=.49\linewidth, clip, trim=0 2.0cm 0 1.0cm]{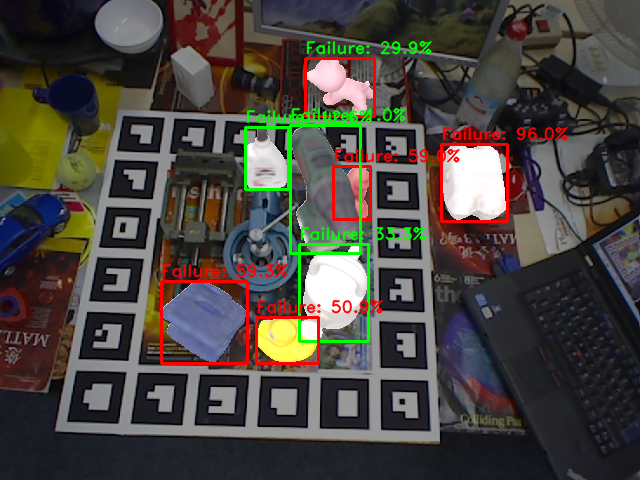}\hfill%
    \includegraphics[width=.49\linewidth, clip, trim=0 1.5cm 0 1.5cm]{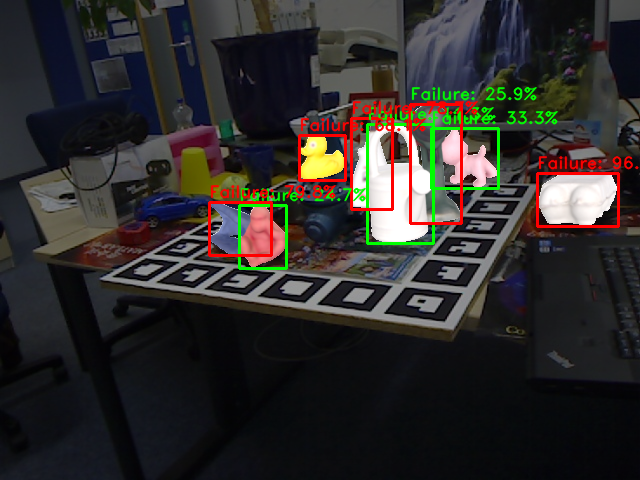}\\[.5em]
    \includegraphics[width=.49\linewidth, clip, trim=0 1.5cm 0 1.5cm]{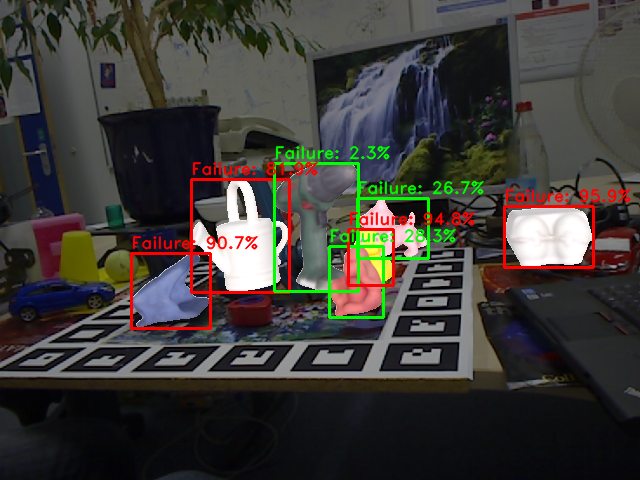}\hfill%
    \includegraphics[width=.49\linewidth, clip, trim=0 1.5cm 0 1.5cm]{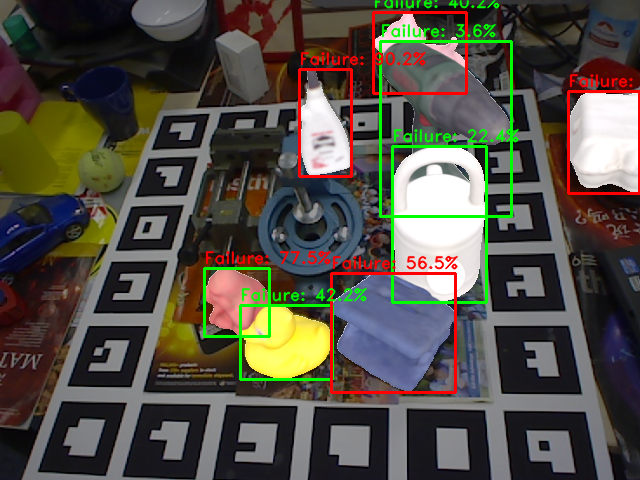}\\[.5em]
    \includegraphics[width=.49\linewidth, clip, trim=0 2.6cm 0 0.4cm]{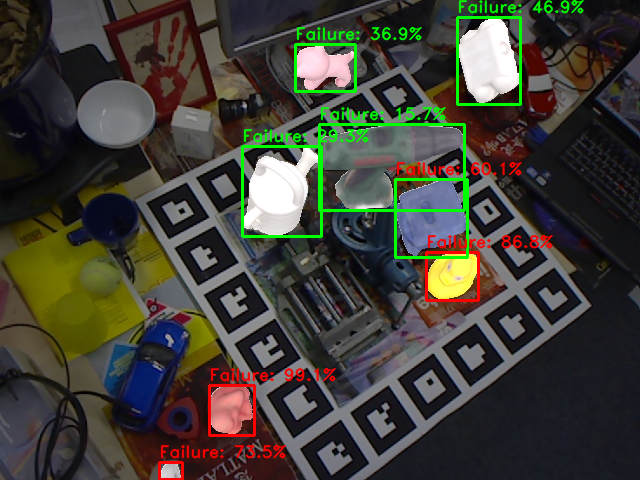}\hfill%
    \includegraphics[width=.49\linewidth, clip, trim=0 1.1cm 0 1.9cm]{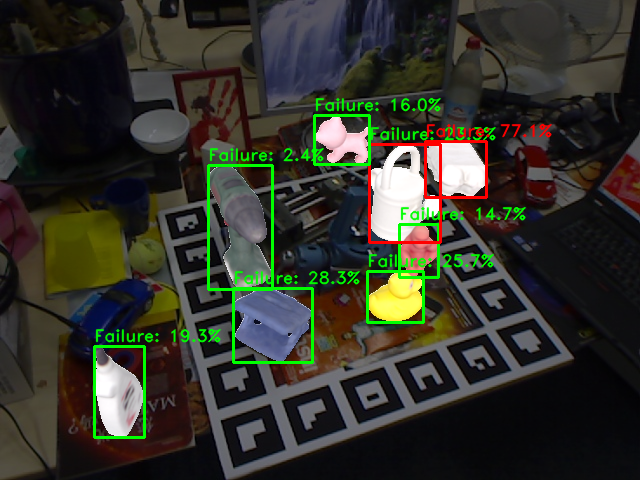}\\[.5em]
    \includegraphics[width=.49\linewidth, clip, trim=0 1.5cm 0 1.5cm]{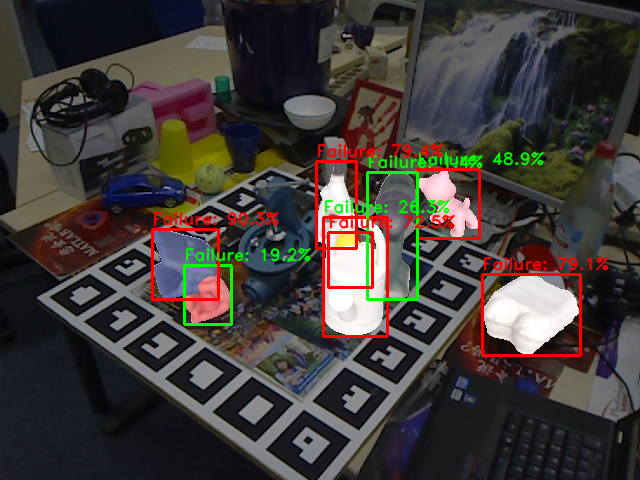}\hfill%
    \includegraphics[width=.49\linewidth, clip, trim=0 1.9cm 0 1.1cm]{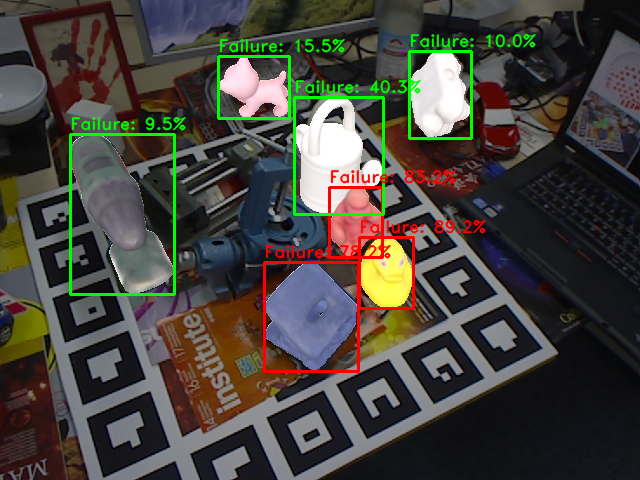}\\[.5em]
    \caption{Pose estimation failure detection $\theta > 5^\circ$ with our Meta Pose approach. The objects are rendered with their estimated poses. The corresponding bounding boxes are color-coded by rotation error: green indicates a correct pose and red an incorrect pose. The predicted failure probability from our failure detection model is annotated on each bounding box.}
    \label{fig:qual-examples}
\end{figure*}

\begin{table*}[p]
    \centering
     \scalebox{.99}{
        \begin{tabular}{c||c|c|c|c|c|c|c|c||c}
            \toprule
            Object & Ape & Can & Cat & Driller & Duck & Eggbox$^*$ & Glue$^*$ & Holepunch &  \\ 
            \midrule\addlinespace[5pt]\midrule
            \multicolumn{10}{c}{Classification rotation angle $\theta > 5^\circ$} \\
            \midrule
            $\varepsilon_{\tau=5}$ & 542 / 1169 & 381 / 1206 & 662 / 1184 & 210 / 1214 & 823 / 1143 & 752 / 1166 & 596 / 900 & 730 / 1210 & Avg \\
            \midrule & \multicolumn{8}{c||}{Maximum Keypoint Confidence Thresholding} & \\ \midrule
            Accuracy & 64.33$\pm$0.00 & 64.43$\pm$0.00 & 62.16$\pm$0.00 & 82.62$\pm$0.00 & 72.00$\pm$0.00 & 65.09$\pm$0.00 & 66.11$\pm$0.00 & 60.33$\pm$0.00& 67.13 \\
            $\text{F}_1$\,score & 59.79$\pm$0.00 & 39.66$\pm$0.00 & 69.40$\pm$0.00 & 37.76$\pm$0.00 & 83.71$\pm$0.00 & 78.45$\pm$0.00 & 79.60$\pm$0.00 & 75.21$\pm$0.00& 65.45 \\
            AUROC & 67.96$\pm$0.00 & 57.05$\pm$0.00 & 70.78$\pm$0.00 & 69.27$\pm$0.00 & 62.26$\pm$0.00 & 71.79$\pm$0.00 & 70.97$\pm$0.00 & 60.01$\pm$0.00& 66.26 \\
            AUPRC & 69.67$\pm$0.00 & 47.10$\pm$0.00 & 79.55$\pm$0.00 & 40.41$\pm$0.00 & 82.53$\pm$0.00 & 84.60$\pm$0.00 & 85.06$\pm$0.00 & 71.08$\pm$0.00& 70.00 \\
            \midrule & \multicolumn{8}{c||}{Conformal Keypoint Prediction} & \\ \midrule
            Accuracy & 74.68$\pm$0.53 & 67.68$\pm$1.62 & 70.54$\pm$0.88 & 85.95$\pm$0.85 & 71.39$\pm$0.00 & 70.09$\pm$0.03 & 75.33$\pm$0.97 & 61.70$\pm$0.40 & 72.17 \\
            $\text{F}_1$\,score & 71.34$\pm$0.80 & 44.53$\pm$2.60 & 73.86$\pm$0.72 & 62.98$\pm$1.38 & 83.31$\pm$0.00 & 82.41$\pm$0.02 & 83.23$\pm$0.05 & 73.45$\pm$0.57& 71.89 \\
            AUROC & 78.98$\pm$0.33 & 62.48$\pm$0.91 & 76.98$\pm$0.13 & 87.30$\pm$0.47 & 54.09$\pm$1.00 & 68.60$\pm$0.76 & 77.01$\pm$1.60 & 63.97$\pm$0.58& 71.18 \\
            AUPRC & 81.06$\pm$0.22 & 54.57$\pm$0.90 & 85.14$\pm$0.16 & 57.96$\pm$1.57 & 77.74$\pm$0.44 & 85.54$\pm$0.41 & 87.01$\pm$0.91 & 71.58$\pm$0.60& 75.08 \\
            \midrule & \multicolumn{8}{c||}{Ours: Meta Pose -- Full} & \\ \midrule
            Accuracy & 76.03$\pm$1.93 & 73.57$\pm$2.10 & 72.83$\pm$0.94 & 83.42$\pm$1.04 & 78.61$\pm$1.21 & 87.39$\pm$0.73 & 79.44$\pm$1.08 & 71.03$\pm$1.13& 77.79 \\
            $\text{F}_1$\,score & 72.69$\pm$1.88 & 56.10$\pm$1.44 & 75.07$\pm$0.64 & 56.85$\pm$1.60 & 85.21$\pm$1.21 & 89.96$\pm$0.65 & 84.70$\pm$1.33 & 76.74$\pm$1.80& 74.66 \\
            AUROC & 82.03$\pm$0.91 & 75.44$\pm$2.21 & 81.05$\pm$0.29 & 84.41$\pm$1.38 & 83.38$\pm$0.71 & 92.36$\pm$0.25 & 86.32$\pm$0.72 & 76.38$\pm$0.82& 82.67 \\
            AUPRC & 83.06$\pm$1.38 & 64.23$\pm$2.38 & 86.85$\pm$0.55 & 57.79$\pm$3.07 & 92.31$\pm$0.60 & 96.07$\pm$0.33 & 92.23$\pm$0.77 & 81.73$\pm$0.90& 81.78 \\
            \midrule & \multicolumn{8}{c||}{Ours: Meta Pose -- Render-free} & \\ \midrule
            Accuracy & 76.52$\pm$1.51 & 73.94$\pm$1.91 & 73.25$\pm$1.85 & 84.62$\pm$0.88 & 79.62$\pm$1.13 & 85.98$\pm$1.03 & 80.42$\pm$1.73 & 71.65$\pm$1.22& 78.25 \\
            $\text{F}_1$\,score & 72.22$\pm$0.85 & 58.74$\pm$2.73 & 76.26$\pm$0.76 & 59.23$\pm$1.54 & 86.36$\pm$1.05 & 88.95$\pm$0.80 & 85.32$\pm$1.79 & 77.76$\pm$1.20& 75.61 \\
            AUROC & 81.94$\pm$1.37 & 77.30$\pm$3.24 & 81.71$\pm$0.77 & 85.87$\pm$0.45 & 84.10$\pm$0.49 & 91.38$\pm$0.57 & 86.64$\pm$1.03 & 75.39$\pm$0.91& 83.04 \\
            AUPRC & 83.07$\pm$1.82 & 66.45$\pm$3.13 & 87.37$\pm$0.70 & 57.18$\pm$1.49 & 92.76$\pm$0.59 & 95.46$\pm$0.45 & 92.24$\pm$0.76 & 80.12$\pm$1.04& 81.83 \\
            \midrule\addlinespace[5pt]\midrule
            \multicolumn{10}{c}{Classification rotation angle $\theta > 10^\circ$} \\
            \midrule
            $\varepsilon_{\tau=10}$ & 227 / 1169 & 70 / 1206 & 335 / 1184 & 43 / 1214 & 338 / 1143 & 400 / 1166 & 329 / 900 & 94 / 1210 & Avg \\
            \midrule & \multicolumn{8}{c||}{Maximum Keypoint Confidence Thresholding} & \\ \midrule
            Accuracy & 85.54$\pm$0.00 & 95.52$\pm$0.00 & 84.71$\pm$0.00 & 96.29$\pm$0.00 & 74.02$\pm$0.00 & 81.48$\pm$0.00 & 79.00$\pm$0.00 & 91.49$\pm$0.00& 86.01 \\
            $\text{F}_1$\,score & 51.85$\pm$0.00 & 55.74$\pm$0.00 & 67.96$\pm$0.00 & 32.84$\pm$0.00 & 47.25$\pm$0.00 & 71.05$\pm$0.00 & 71.15$\pm$0.00 & 28.97$\pm$0.00& 53.35 \\
            AUROC & 79.28$\pm$0.00 & 83.67$\pm$0.00 & 84.30$\pm$0.00 & 79.56$\pm$0.00 & 70.05$\pm$0.00 & 84.57$\pm$0.00 & 84.66$\pm$0.00 & 70.90$\pm$0.00& 79.62 \\
            AUPRC & 61.93$\pm$0.00 & 57.90$\pm$0.00 & 78.04$\pm$0.00 & 30.69$\pm$0.00 & 59.67$\pm$0.00 & 81.89$\pm$0.00 & 80.61$\pm$0.00 & 29.12$\pm$0.00& 59.98 \\
            \midrule & \multicolumn{8}{c||}{Conformal Keypoint Prediction} & \\ \midrule
            Accuracy & 86.09$\pm$0.34 & 97.05$\pm$0.17 & 86.47$\pm$0.74 & 94.96$\pm$0.72 & 75.77$\pm$0.79 & 72.86$\pm$0.59 & 75.64$\pm$1.01 & 91.31$\pm$1.42 & 85.02 \\
            $\text{F}_1$\,score & 63.68$\pm$0.46 & 66.05$\pm$1.36 & 76.04$\pm$0.74 & 31.49$\pm$2.00 & 43.27$\pm$0.98 & 50.86$\pm$0.62 & 65.89$\pm$2.58 & 44.46$\pm$2.04& 55.22 \\
            AUROC & 88.53$\pm$0.46 & 91.69$\pm$0.35 & 91.88$\pm$0.25 & 91.46$\pm$0.37 & 63.19$\pm$1.10 & 63.44$\pm$0.45 & 82.27$\pm$1.81 & 82.98$\pm$0.84& 81.93 \\
            AUPRC & 71.87$\pm$0.47 & 70.60$\pm$1.37 & 84.35$\pm$0.41 & 31.18$\pm$1.07 & 56.73$\pm$0.72 & 59.01$\pm$0.15 & 74.01$\pm$1.44 & 41.99$\pm$1.57& 61.22 \\
            \midrule & \multicolumn{8}{c||}{Ours: Meta Pose -- Full} & \\ \midrule
            Accuracy & 92.45$\pm$0.83 & 96.97$\pm$0.34 & 87.90$\pm$1.37 & 95.12$\pm$0.77 & 76.20$\pm$2.13 & 88.74$\pm$0.79 & 84.94$\pm$0.56 & 89.79$\pm$2.68& 89.02 \\
            $\text{F}_1$\,score & 81.16$\pm$1.31 & 74.91$\pm$3.46 & 78.86$\pm$2.11 & 45.34$\pm$1.79 & 60.28$\pm$2.98 & 83.25$\pm$1.67 & 79.97$\pm$1.97 & 45.75$\pm$7.47& 68.69 \\
            AUROC & 95.96$\pm$0.68 & 98.11$\pm$0.32 & 94.59$\pm$0.65 & 87.78$\pm$5.20 & 80.78$\pm$2.83 & 95.80$\pm$0.24 & 93.32$\pm$0.62 & 81.74$\pm$3.57& 91.01 \\
            AUPRC & 88.85$\pm$0.96 & 84.60$\pm$1.66 & 88.90$\pm$1.20 & 40.83$\pm$4.30 & 68.74$\pm$2.38 & 92.98$\pm$0.29 & 88.97$\pm$0.67 & 48.06$\pm$8.44& 75.24 \\
            \midrule & \multicolumn{8}{c||}{Ours: Meta Pose -- Render-free} & \\ \midrule
            Accuracy & 92.58$\pm$0.79 & 96.89$\pm$0.58 & 87.69$\pm$1.44 & 96.17$\pm$0.54 & 75.83$\pm$0.81 & 87.39$\pm$0.66 & 84.86$\pm$0.91 & 91.22$\pm$1.54& 89.08 \\
            $\text{F}_1$\,score & 81.60$\pm$1.35 & 71.67$\pm$8.07 & 78.37$\pm$2.55 & 49.43$\pm$3.11 & 59.77$\pm$4.94 & 81.02$\pm$1.25 & 79.55$\pm$2.32 & 45.11$\pm$7.25& 68.31 \\
            AUROC & 95.99$\pm$0.64 & 94.34$\pm$7.88 & 94.81$\pm$0.59 & 89.74$\pm$4.61 & 80.11$\pm$2.33 & 94.76$\pm$0.60 & 93.24$\pm$0.46 & 81.02$\pm$4.23& 90.50 \\
            AUPRC & 88.91$\pm$0.78 & 80.29$\pm$9.11 & 89.16$\pm$1.01 & 46.09$\pm$1.32 & 68.18$\pm$2.10 & 91.19$\pm$0.53 & 88.78$\pm$0.75 & 45.98$\pm$8.29& 74.82 \\
            \bottomrule
        \end{tabular}
    }\vspace{1em}
        \caption{Class-wise detection performance of pose estimation failures on LINEMOD Occluded \cite{hodan2018bop}. We compare maximum keypoint confidence and conformal keypoint to our approach Meta Pose. Here, $^*$ denotes symmetric objects, and $\varepsilon_\tau$ the pose error rate at threshold angle $\tau$ with the underlying keypoint localization model by \citet{schmeckpeper2022semantic}.}
        \label{tab:results}
\end{table*}

\textbf{Per-Object Failure Detection.} \Cref{tab:results} reports the per-object failure detection performance across all eight LINEMOD Occluded objects \cite{hodan2018bop} for both rotation error thresholds $\tau = 5^\circ$ and $\tau = 10^\circ$, comparing maximum keypoint confidence thresholding \cite{hendrycks2017baseline}, conformal keypoint prediction \cite{yang2023conformal}, and our proposed Meta Pose approaches. 

For $\theta > 5^\circ$ Meta Pose achieves the highest average AUROC of 83.04 and AUPRC of 81.83, outperforming conformal keypoint prediction (71.18 and 75.08) and confidence thresholding (66.26 and 70.00) by a substantial margin. The gains are particularly pronounced for the symmetric objects \textbf{eggbox} and \textbf{glue}, reaching AUROC values of 91.38 and 86.64 respectively, and AUPRC values of 95.46 and 92.24. The result for the object \textbf{duck} is similar high, achieving an AUPRC of 92.76. Objects with low pose error rates such as \textbf{driller} (210 failures out of 1214 samples) present a harder failure detection problem due to severe class imbalance, which can be observed by the lower F$_1$ scores across all methods.

For $\theta > 10^\circ$ class imbalance becomes even more pronounced, \eg the object \textbf{driller} has only 43 failures. Still, Meta Pose maintains strong AUROC performance, reaching an average of 90.50 compared to 81.93 for conformal keypoint prediction and 79.62 for maximum keypoint confidence thresholding. Notably, for the objects \textbf{ape}, \textbf{can}, \textbf{cat}, \textbf{eggbox} and \textbf{glue}, Meta Pose achieves AUROC values over 90 percent points, respectively, demonstrating that the proposed consistency features become increasingly discriminative despite failures becoming rarer. This indicates object specific failure modes which can be detected more reliably.

Furthermore, the consistently small standard deviations across all metrics and objects indicate that the failure detector is robust to the choice of calibration fold in the reverse cross-validation protocol, suggesting that even a small labeled subset from the target domain is sufficient to train a stable and reliable failure detection model.

Comparing Meta Pose--Full and its render-free variant Meta Pose, we observe similar performance on any metric.
This indicates that the additional render- and mask-based features provide no clear benefit. For $\theta > 10^\circ$, Meta Pose--Full reaches an average AUROC of 91.01 and AUPRC of 75.24, compared to 90.50 AUROC and 74.82 AUPRC for render-free variant, which is the only marginal gain across metrics. 
We consider the render-free variant as preferred choice in practice due to its lower computational costs, \cf \Cref{sec:ablation}. Both variants consistently and substantially outperform conformal keypoint prediction and confidence thresholding across nearly all objects and both thresholds, confirming that spatial self-consistency of keypoint predictions provides a reliable signal for failure detection independent of render-based features.\\

\begin{figure}
    \centering
    \includegraphics[width=.99\linewidth]{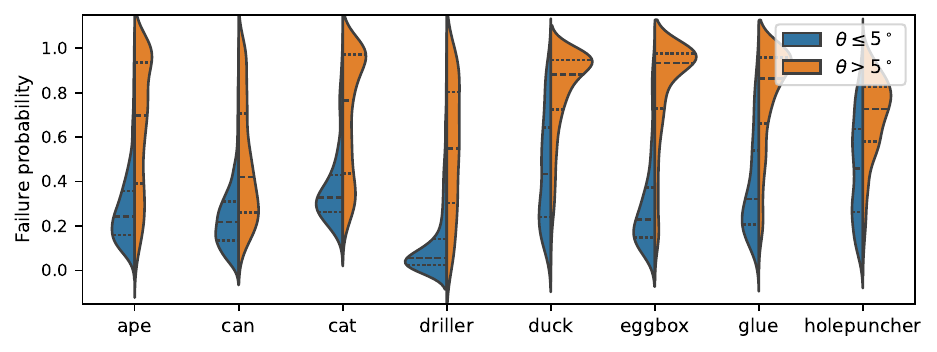}\\%
    \includegraphics[width=.99\linewidth]{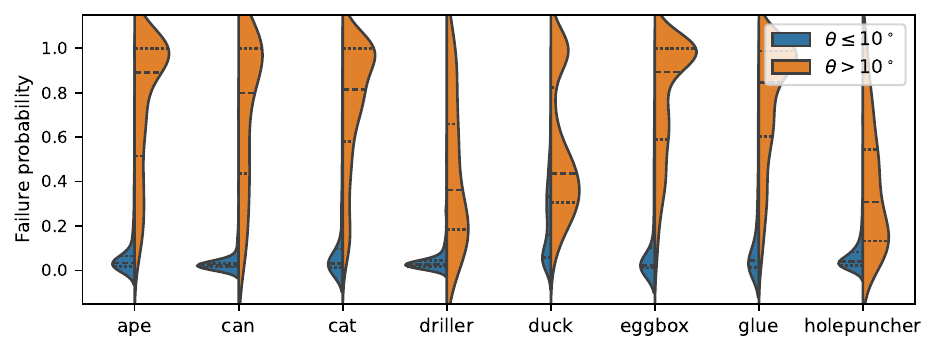}
    \caption{Violin plots for our Meta Pose failure detection model in LINEMOD Occluded test \cite{hodan2018bop}. From left to right: failure detection for error threshold at $\tau=5^\circ$ and $\tau=10^\circ$}
    \label{fig:violins}
\end{figure}

\textbf{Failure Probability Distributions.}
The violin plots in \Cref{fig:violins} support the claim that incorrect pose estimates tend to receive higher predicted failure probabilities, though the degree of separation varies across objects but is consistent with the per-object AUROC values reported in \Cref{tab:results}. \textbf{Eggbox} and \textbf{glue} exhibit the clearest separation, with correct poses ($\theta \leq \tau$) concentrated at low failure probabilities and incorrect poses ($\theta > \tau$) massed near one, indicating confident and accurate classifier predictions. For \textbf{driller} and \textbf{duck}, the predicted probabilities are more spread across the full range for one class, yet the opposing class remains tightly concentrated, which preserves good overall separability. \textbf{Ape} and \textbf{cat} follow a similar pattern, though with generally higher predicted probabilities across both classes and correspondingly shorter distribution tails, suggesting a systematic tendency of the classifier to predict failures for these objects. For \textbf{can} and \textbf{holepuncher}, incorrect poses still receive higher failure probabilities than correct ones, as evidenced by the relative positions of the distribution modes, though the substantial overlap between the two distributions leads to a higher rate of false positives or false negatives depending on the chosen decision threshold. Across all objects, the consistent shift of the incorrect pose 
distribution towards higher predicted probabilities confirms that the geometric consistency features capture meaningful information about pose estimation failures.\\

\begin{figure}[t]
    \includegraphics[width=.49\linewidth]{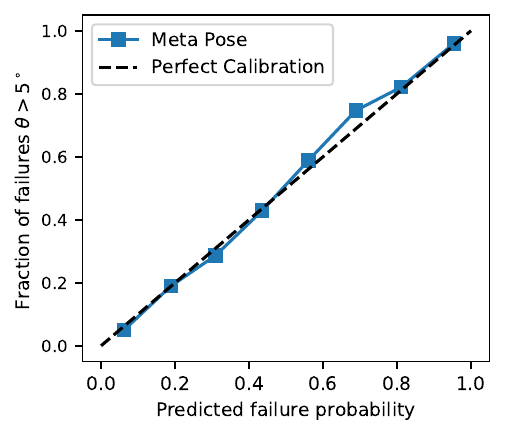}%
    \includegraphics[width=.49\linewidth]{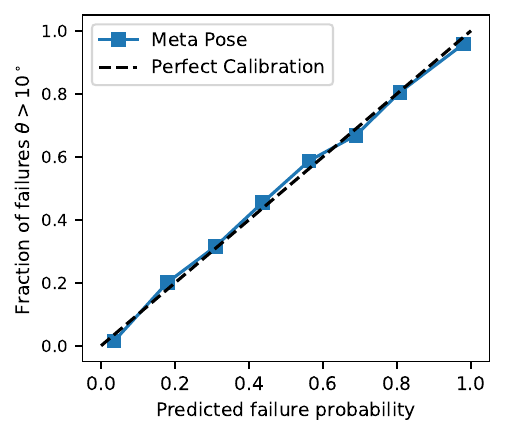}
    \caption{Calibration curves for both rotation error thresholds $\tau = 5^\circ$ and $\tau = 10^\circ$ of the logistic regression model within our Meta Pose approach.}
    \label{fig:calibration}
\end{figure}

\textbf{Calibration.}
\Cref{fig:calibration} shows the calibration curves of our failure detection model for both rotation error thresholds. A perfectly calibrated model follows the diagonal line, meaning that a predicted failure probability $\hat{p}$ corresponds to an empirical failure frequency. A well calibrated failure detection model is particularly important in our setting, as it ensures that $\hat{p}$ can be interpreted as a meaningful confidence score for a failure, which is essential for reliable downstream decisions such as triggering pose re-estimation or flagging samples.

For $\tau = 5^\circ$, the logistic regression model is well calibrated across the full probability range with only a slight tendency towards overconfidence. For $\tau = 10^\circ$, the calibration is even stronger with the calibration curve closely following the diagonal. Both curves remain close to the diagonal overall, demonstrating that the logistic regression model produces well-calibrated failure probabilities across both rotation error thresholds.

\begin{figure}[]
    \includegraphics[width=.49\linewidth]{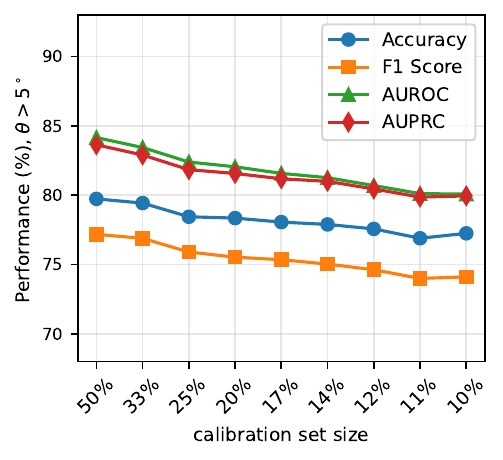}%
    \includegraphics[width=.49\linewidth]{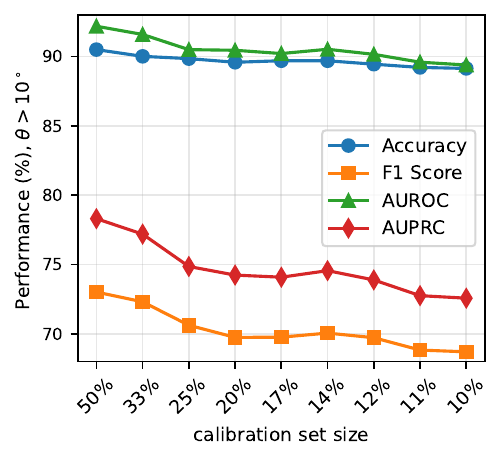}
    \caption{Sample efficiency of our proposed Meta Pose failure detection model.
    The evaluation metrics are reported as a function of the calibration set size. }
    \label{fig:sample_efficieny}
\end{figure}

\section{Ablation Study}\label{sec:ablation}\vspace{1em}

\textbf{Sample Efficiency.} \Cref{fig:sample_efficieny} shows the failure detection performance as a function of the calibration set size, ranging from 50\% down to 10\% of the available data. For error threshold $\tau = 5^\circ$, all four metrics remain stable across the full range of calibration set sizes. The overall performance drop from 50\% down to 10\% is small, demonstrating that the classifier is not sensitive to the amount of calibration data in this setting.

For error threshold $\tau = 10^\circ$, AUPRC and F$_1$ score show a more noticeable decline as the calibration set shrinks, which can be attributed to the more severe class imbalance at this threshold, \cf pose error rates in \Cref{tab:results} which for instance for the objects \textbf{can}, \textbf{driller}, \textbf{holepuncher} are less than 10\%. The number of failures per object drops substantially the smaller the calibration set size, making it more difficult to learn a reliable failure detector from fewer labeled samples. Nevertheless, even in this more challenging setting the degradation remains moderate, with AUPRC and F$_1$ losing $\sim5$ percent points, indicating that the failure detection has meaningful discriminative power. All together, both plots confirm that only a small calibration set drawn from the target domain is sufficient to train a reliable failure detector.\\

\begin{table}[t]
\centering
\setlength{\tabcolsep}{2.5pt}
\renewcommand{\arraystretch}{1.05}

\begin{tabular}{ccccccc cccc}
\toprule
\multicolumn{3}{c}{keypoints} & \multicolumn{2}{c}{render} & \multicolumn{2}{c}{mask} & \multicolumn{4}{c}{$\theta > 5^\circ$} \\
\cmidrule(lr){1-3} \cmidrule(lr){4-5} \cmidrule(lr){6-7} \cmidrule(lr){8-11}
$\mathbf{d}^\text{rel}$ & $\mathbf{d}^\text{repr}$ & $\mathbf{u}$ & $\mathbf{d}^\text{render}$ & $\mathbf{d}^\text{center}$ & $d^\text{cham}$ & $r^\text{cov}_{\delta=2}$ & Acc & F$_1$ & AUROC & AUPRC \\
\midrule
\checkmark & - & - & - & - & - & - &74.73 & 71.70 & 78.61 & 77.91 \\
 - & \checkmark & - & - & - & - & - & 75.22 & 71.54 & 78.45 & 79.14 \\
 - & - & \checkmark & - & - & - & - & 75.34 & 73.01 & 79.81 & 80.03 \\
 - & \checkmark & \checkmark & - & - & - & - &76.32 & 73.21 & 80.40 & 80.46 \\
\checkmark & \checkmark & \checkmark & - & - & - & - & \textbf{78.25} & \textbf{75.61} & \textbf{83.04} & \textbf{81.83} \\
\midrule
-& - & - &  \checkmark & \checkmark & - & - & 71.19 & 68.75 & 73.22 & 73.27 \\
 -& - & - & - & - & \checkmark & \checkmark & 74.54 & 71.97 & 77.65 & 77.53 \\
 -& - & - & \checkmark & \checkmark & \checkmark & \checkmark & 74.66 & 71.74 & 77.83 & 77.84 \\
\midrule
\checkmark & \checkmark & \checkmark & \checkmark & \checkmark & \checkmark & \checkmark & 77.79 & 74.66 & 82.67 & 81.78 \\
\bottomrule
\end{tabular}

\vspace{0.8em}

\begin{tabular}{ccccccc cccc}
\toprule
\multicolumn{3}{c}{keypoints} & \multicolumn{2}{c}{render} & \multicolumn{2}{c}{mask} & \multicolumn{4}{c}{$\theta > 10^\circ$} \\
\cmidrule(lr){1-3} \cmidrule(lr){4-5} \cmidrule(lr){6-7} \cmidrule(lr){8-11}
$\mathbf{d}^\text{rel}$ & $\mathbf{d}^\text{repr}$ & $\mathbf{u}$ & $\mathbf{d}^\text{render}$ & $\mathbf{d}^\text{center}$ & $d^\text{cham}$ & $r^\text{cov}_{\delta=2}$ & Acc & F$_1$ & AUROC & AUPRC \\
\midrule
\checkmark & - & - & - & - & - & - & 85.17 & 53.97 & 79.84 & 59.28 \\
 - & \checkmark & - & - & - & - & - & 88.29 & 66.95 & 88.76 & 73.43 \\
 - & - & \checkmark & - & - & - & - & 87.26 & 62.33 & 88.47 & 69.35 \\
 - & \checkmark & \checkmark & - & - & - & - & 88.56 & 67.66 & 89.77 & 73.98 \\
\checkmark & \checkmark & \checkmark & - & - & - & - & \textbf{89.08} & 68.31 & 90.50 & 74.82 \\
\midrule
 -& - & - &  \checkmark & \checkmark & - & - & 83.77 & 52.81 & 79.67 & 55.51 \\
 -& - & - & - & - & \checkmark & \checkmark & 85.55 & 56.64 & 86.19 & 63.79 \\
 -& - & - & \checkmark & \checkmark & \checkmark & \checkmark & 85.65 & 59.72 & 85.99 & 65.45 \\
\midrule
\checkmark & \checkmark & \checkmark & \checkmark & \checkmark & \checkmark & \checkmark & 89.02 & \textbf{68.69} & \textbf{91.01} & \textbf{75.24} \\
\bottomrule
\end{tabular}

\caption{Failure detection performance of different combinations of features for $\theta > 5^\circ$ (top) and $\theta > 10^\circ$ (bottom). Render-free refers to $\{\mathbf{d}^\text{rel}, \mathbf{d}^\text{repr}, \mathbf{u}\}$. Full additionally includes $\{\mathbf{d}^\text{render}, \mathbf{d}^\text{center}, d^\text{cham}, r^\text{cov}_{\delta=2} \}$.}
\label{tab:ablation-features}

\end{table}

\textbf{Feature Ablation.} In \Cref{tab:ablation-features}, we report the performance of failure detection for different combinations of features, evaluated separately for $\theta > 5^\circ$ and $\theta > 10^\circ$. Across both thresholds, we observe that aggregating features consistently improves discriminative performance over any individual feature, suggesting that pose estimation failures result from multiple sources of geometric inconsistency rather than a single dominant signal.

Moreover, we observe that combining all non-render features yields the strongest performance among individual and paired feature combinations. Render-based and mask-based features in isolation, and also combined with each other, are consistently outperformed by the non-render combination. For $\theta > 5^\circ$, the combination of all features does not improve over Meta Pose (which only uses the render-free features) on any metric. For $\theta > 10^\circ$, the combination of all features yields marginal improvements in $\mathrm{F}_1$, AUROC, and AUPRC of at most $0.51$ percent points across all three metrics, while Meta Pose remains best in accuracy. Given the additional runtime cost of rendering and segmentation required for the render- and mask-based features, cf. \Cref{tab:ablation-runtime}, we conclude that these marginal gains do not to outweigh the added computational overhead, and therefore adopt the render-free Meta Pose as our main approach of this work.\\

\begin{table}[t]
\centering
\begin{tabular}{lcc}
\toprule
Method & Runtime (ms / image) & Relative \\
\midrule
Max Keypoint Confidence   & 330  & $1.0\times$ \\
Conformal Keypoint        & 2030 & $6.2\times$ \\
Ours: Meta Pose -- Full   & 1220 & $3.7\times$ \\
Ours: Meta Pose (render-free) & 374  & $1.1\times$ \\
\bottomrule
\end{tabular}
\caption{Runtime comparison of the different failure detection methods. Relative runtime is given with respect to the Maximum Keypoint Confidence baseline, \cf \Cref{sec:setup}.}
\label{tab:ablation-runtime}
\end{table}

\textbf{Runtime Comparison.} \Cref{tab:ablation-runtime} reports the runtime per image for each method. Meta Pose (render-free) adds only little overhead compared to the maximum keypoint confidence with 374 ms per image. The geometric consistency features are computationally lightweight by design. They require only pairwise distance computations between different keypoint combinations, and reprojections between 2D and 3D are simple matrix multiplications that can be done efficiently. 

Conformal keypoint prediction, by contrast, is roughly six times slower at 2030 ms per image. This overhead stems from two sources. First, computing inductive conformal prediction scores requires processing the full keypoint localization heatmaps rather than simply collecting statistics from keypoints obtained from a single forward pass of the pose estimation pipeline. Second, uncertainty propagation is performed by sampling a set of possible poses from the conformal keypoint sets, which in our experiments uses a sample size of 50 poses. 

Meta Pose--Full, which additionally includes the render- and mask-based features, requires 1220 ms per image, roughly $3.7\times$ the baseline. Given that Meta Pose--Full offers only marginal performance gains over the render-free variant, cf. \Cref{tab:ablation-features}, this runtime cost further supports our choice of the render-free Meta Pose variant as the main approach. Together, these results demonstrate that Meta Pose achieves reliable failure detection performance at low computational cost, making it well suited for integration into real-time or resource-constrained pose estimation pipelines.\\

\begin{figure}[t]
    \includegraphics[width=.49\linewidth]{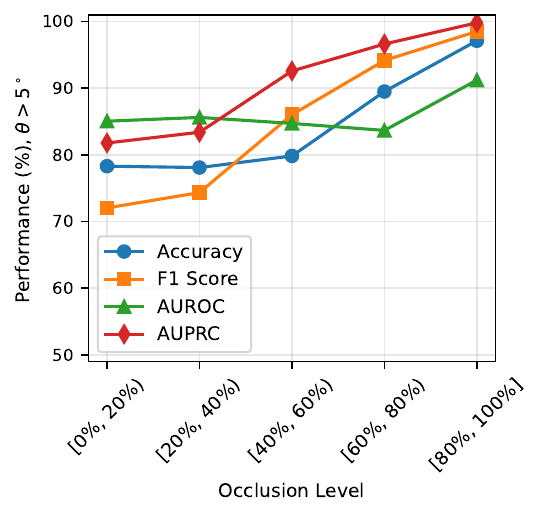}%
    \includegraphics[width=.49\linewidth]{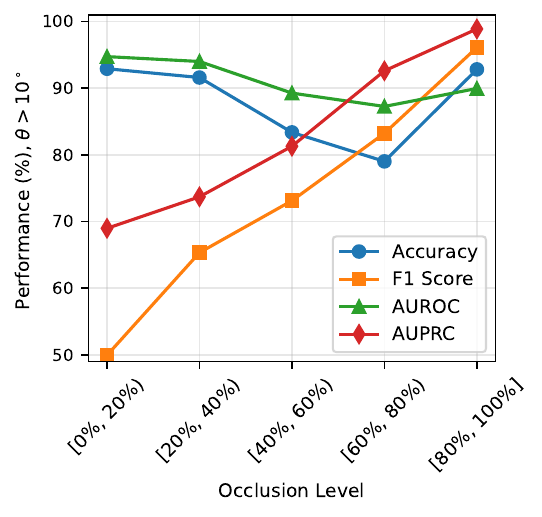}
    \caption{Failure detection performance of our Meta Pose failure detection model at different occlusion levels of objects.}
    \label{fig:performance_occlusion}
\end{figure}

\begin{table}[t]
    \centering
    \vspace{2em}
    \begin{tabular}{|c|c|c|c|c|c|}
        \hline
        Occlusion & $[0,20)$ & $[20,40)$ & $[40,60)$ & $[60,80)$ & $[80,100]$ \\
        \hline\hline
        \# samples & 5478 & 1658 & 1017 & 620 & 419 \\
        \hline\hline
        $\varepsilon_{\tau=5}$ & 0.42 & 0.43 &  0.70 & 0.87 & 0.98 \\
        $\varepsilon_{\tau=10}$ & 0.09 & 0.13 & 0.32 & 0.63 & 0.91 \\
        \hline
    \end{tabular}
    \caption{Pose error rates, \cf \Cref{eq:pose-error}, at different occlusion levels of objects.}
    \label{tab:occlusion_pose_error_rate}
\end{table}

\textbf{Failure Detection vs.\ Occlusion Level.}~ 
\Cref{fig:performance_occlusion} reports the failure detection performance as a function of occlusion levels for both rotation error thresholds. Since pose estimation becomes increasingly challenging as occlusion increases, reliable failure detection is particularly important in highly occluded scenes. Both plots show a consistent trend. Failure detection performance improves with occlusion level in AUPRC and F1 score, reaching near-perfect performance at occlusion levels above $80\%$. This is an encouraging result, as it demonstrates that the proposed method is most reliable in scenarios where pose estimation is most likely to fail.

For $\tau = 5^\circ$, accuracy, F1 score and AUPRC even show monotonic improvement with increasing occlusion, while AUROC remains stable but still with best performance at highest occlusion. This suggests that the failure detector becomes more decisive as occlusion increases. 

For $\tau = 10^\circ$, a more pronounced improvement is observed across all metrics, with F1 score increasing from $50\%$ at low occlusion to above $95\%$ at high occlusion. The low F1 score at low occlusion for $\tau = 10^\circ$ can be explained by the severe class imbalance, \cf \Cref{tab:occlusion_pose_error_rate}. When occlusion is low, pose estimation rarely fails for $\tau = 10^\circ$, leaving very few samples with failures for the classifier to detect. Moreover, class imbalance drives the metrics accuracy and AUROC. 

So the dip in accuracy and AUROC rather reflects the disappearance of the severe class imbalance rather than a deterioration of the failure detection model. Importantly, all metrics remain clearly above their respective no-skill baselines, \ie $(1 - \varepsilon_t)$ for accuracy and $50\%$ for AUROC, confirming that the classifier has meaningful discriminative capabilities across all occlusion levels. 

For a complete assessment of failure detection performance, all metrics must be taken into account. Since F1 score and AUPRC measure failure detection performance independently of the class distribution, their consistent improvement with occlusion level supports that the proposed failure detection model becomes more reliable as occlusion increases.\\

\section{Conclusion and Outlook}\vspace{1em}

In this work, we presented Meta Pose, a lightweight framework for detecting pose estimation failures in keypoint-based 6D pose estimation. Our approach exploits the geometric self-consistency of predicted 2D keypoints across complementary sources, including the input image, reprojections under the estimated pose, and keypoint and mask predictions on the rendered object. We construct a set of hand-crafted features that serve as input to a logistic regression classifier. Despite its simplicity, Meta Pose consistently outperforms baselines, including maximum keypoint confidence thresholding and conformal keypoint prediction, across all eight objects of the LINEMOD Occluded dataset and at multiple rotation error thresholds. We additionally introduced Meta Pose-Full, which augments our render-free feature set with render- and mask-based consistency features, and showed through an ablation study that these additional features yield only marginal performance gains at substantially higher computational cost. We therefore adopt the render-free Meta Pose as our main approach, which runs at low computational overhead relative to the standard pose estimation pipeline, requires only a small labeled calibration set from the target domain, and produces well-calibrated failure probabilities.

There are directions which remain open for future work. Meta Pose--Full relies on the Segment Anything Model for its mask consistency features, so its performance is coupled to the quality of this external segmentation model. Improving the segmentation quality, \eg through better prompting, would likely benefit not only failure detection performance but could also be used directly to improve the accuracy of pose estimation pipeline. Moreover, while we focused on rotation error as the primary failure criterion, extending the framework to jointly detect translation failures, \eg uncertainty quantification for bounding box regression, could further improve its utility in downstream tasks such as robotic manipulation. Additionally, the current evaluation is limited to the LINEMOD Occluded dataset. Validating the approach on a broader set of benchmarks, object categories, and pose estimation backbones remains an open direction and would strengthen the generality of our findings. Finally, the predicted failure probabilities could be integrated into the pose estimation pipeline itself, for instance by triggering re-estimation if a failure is likely, thereby closing the loop between failure detection and pose estimation.

\pagebreak
\bibliography{references} 
\bibliographystyle{IEEEtranN}

\end{document}